\documentclass[letterpaper, 10 pt, conference]{ieeeconf}  

\IEEEoverridecommandlockouts                              

\usepackage{color} 

\usepackage{mathrsfs}
\usepackage{cite}
\usepackage{amsmath,amssymb,amsfonts}
\usepackage{algorithmic}
\usepackage{graphicx}
\usepackage{algorithm}
\usepackage{subfigure}
\usepackage{textcomp}
\usepackage{xcolor}
\usepackage{diagbox}
\usepackage{color}
\usepackage{caption}
\usepackage{graphicx}
\usepackage{float}
\usepackage{url}

\title{\LARGE \bf
A Human Motion Compensation Framework for \\ a  Supernumerary Robotic Arm 
}

\author{Xin Zhang$^{1,2,3}$, Pietro Balatti$^{3}$, Mattia Leonori$^{3}$, Arash Ajoudani$^{3}$  
\thanks{*This work was supported in part by the European Union’s Horizon 2020 research and innovation programme SOPHIA (Grant No.871237), in part by the ERC-StG Ergo-Lean (Grant No.850932 ), in part by the National Natural Science Foundation of China (Grant No.62103407), and in part by the China Scholarship Council,}
\thanks{$^{1}$State Key Laboratory of Robotics, Shenyang Institute of Automation, Chinese Academy of Sciences, 110016, Shenyang, China. Email: \tt\small{zhangxin@sia.cn}}
\thanks{$^{2}$Institutes for Robotics and Intelligent Manufacturing, Chinese Academy of Sciences, 110169, Shenyang, China.}
\thanks{$^{3}$Human-Robot Interfaces and Interaction (HRI$^2$) Lab, Istituto Italiano di Tecnologia, 16163 Genova, Italy.  Email: \tt\small{Xin.Zhang@iit.it, Mattia.Leonori@iit.it, Pietro.Balatti@iit.it,  Arash.Ajoudani@iit.it}}
\thanks{This work was carried out at the HRI$^2$ lab.}
\thanks{Corresponding author: Xin Zhang.}
}

\begin{document}

\maketitle
\thispagestyle{empty}
\pagestyle{empty}

\begin{abstract}
Supernumerary robotic arms (SRAs) can be used as the third arm to complement and augment the abilities of human users. The user carrying a SRA forms a connected kinodynamic chain, which can be viewed as a special class of floating-base robot systems. However, unlike general floating-base robot systems, human users are the bases of SRAs and they have their subjective behaviors/motions. This implies that human body motions can unintentionally affect the SRA's end-effector movements. To address this challenge, we propose a framework to compensate for the human whole-body motions that interfere with the SRA's end-effector trajectories.  The SRA system in this study consists of a 6-degree-of-freedom lightweight arm and a wearable interface. The wearable interface allows users to adjust the installation position of the SRA to fit different body shapes. An inertial measurement unit (IMU)-based sensory interface can provide the body skeleton motion feedback of the human user in real time. By simplifying the floating-base kinematics model, we design an effective motion planner by reconstructing the Jacobian matrix of the SRA. Under the proposed framework, the performance of the reconstructed Jacobian method is assessed by comparing the results obtained with the classical nullspace-based method through two sets of experiments. 
\end{abstract}


\section{INTRODUCTION}

Supernumerary robotic limbs (SRLs) provide additional robotic limbs to complement and augment the physical dexterity and capabilities of their human users. Over the past decades, various SRLs (e.g., legs \cite{parietti2015design}, arms \cite{muehlhaus2023need}, and fingers \cite{hussain2017toward}) have been developed for assisting humans to achieve some supporting, grasping, and manipulation tasks. Furthermore, SRLs can reduce risk by minimizing human users' exposure to potential hazards. For instance, they can be used to handle dangerous materials or carry out tasks in sub-areas that may be inaccessible or risky for human operators. Developing intelligent SRLs has attracted lots of attention in the robotics field.

\begin{figure}[htb]
\centerline{\includegraphics[width=0.27\textwidth]{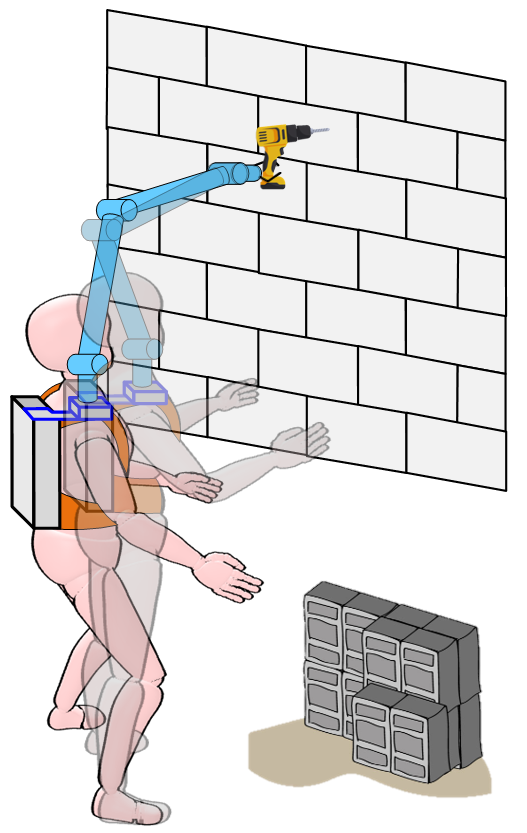}}
\caption{Concept illustration of human motion compensation: when the user body moves, the SRA compensates for the induced motion and decouples it from the end-effector trajectories.}
\label{fig:conception}
\end{figure}


As one of the most representative SRLs, supernumerary robotic arms (SRAs) can seamlessly cooperate with users regardless of work location. The pioneering work \cite{armharry,parietti2014supernumerary} developed two-arm SRA prototypes that can share the static workload acting on human arms in an overhead assembly task. For the human-robot collaboration, the kinematic chains of SRAs are not independent from human limbs, so safe human-robot interfaces and collaborative control frameworks are important aspects that should be taken into consideration. Veronneau et al. \cite{veronneau2020multifunctional} designed a 3-degree-of-freedom (DoF) SRA with a teleoperation control system for agricultural harvest tasks. Saraiji et al. \cite{saraiji2018metaarms} proposed two anthropomorphic SRAs and developed a teleoperation interface mapping the users' feet and leg motion to control the robotic arms. Penaloza et al. \cite{penaloza2018towards} use a non-invasive brain-machine interface to detect the user's intention and control the SRA. Khoramshahi et al. \cite{khoramshahi2023practical} proposed a collision-avoidance algorithm for a 7-DoF SRA using the quadratic programming approach. A nullspace-based controller is proposed to realize collision avoidance by maximizing the distance between the human's head and the robot in our previous work \cite{Du2023Bi}. 

The human-SRA systems can be considered a special class of floating-base robot systems, and their motion depends not only on the system itself but also on the movement of the human user. However, unlike general robotic floating-base systems, the movement of humans is not completely controllable or predictable, which poses a unique challenge for the integrated human-SRA system. For example, while carrying out delicate and/or constrained manipulation tasks or inspecting an external environment, the spontaneous or unintentional movements of human users will disturb the operational accuracy of the SRA. Therefore, the SRA system should be able to compensate for the unpredictable or unintentional motions induced by the users. The conception of the human motion compensation is shown in Fig. \ref{fig:conception}. This also means that the SRA needs to estimate/monitor
the human body state to reduce the induced motion. In this context, we integrate an inertial measurement unit (IMU)-based sensory interface into our human-SRA system and develop a motion compensation control framework to address the above issues. To the best of our knowledge, this paper is the first work on the motion compensation control between human users and SRAs. The main contributions of this work can be summarized as:
\begin{itemize}
\item A physical and kinematic integration of the human-robot system is developed by combining our SRA and an IMU-based sensory interface.
\item A motion compensation control framework is proposed to compensate for the 3-dimensional (3D) translation motion induced by human users.
\item Based on a simplified floating-base kinematics model, an effective motion planner is designed by reconstructing the Jacobian matrix.
\end{itemize}

We focus on the translational axes even if the theory is general and includes the rotational axes. However, since most SRAs have a limited number of DoFs, the isolation of all translational and rotational axes would not be feasible. The rest of the paper is organized as follows. In Section \ref{System}, the human-robot integration system consisting of our SRA and the IMU-based sensory interface is introduced. The floating-base kinematics modeling and analysis are described in Section \ref{theory}. The motion compensation control framework is described in Section \ref{framework}. The experiments and results are illustrated in Section \ref{experiment}. Finally, Section \ref{conclusion} addresses the conclusion of the paper. 

\section{HUMAN-ROBOT INTEGRATION SYSTEM}\label{System}

In this section, we will introduce the design philosophy of our human-SRA system, where the human-robot integration is twofold. The wearable interface keeps the user and the SRA integrated physically. The IMU-based sensory interface works as a bridge between the user and the SRA guaranteeing the kinematic chain integration.

\subsection{Supernumerary Robotic Arm}
\begin{figure}[htb]
\centerline{\includegraphics[width=0.4\textwidth]{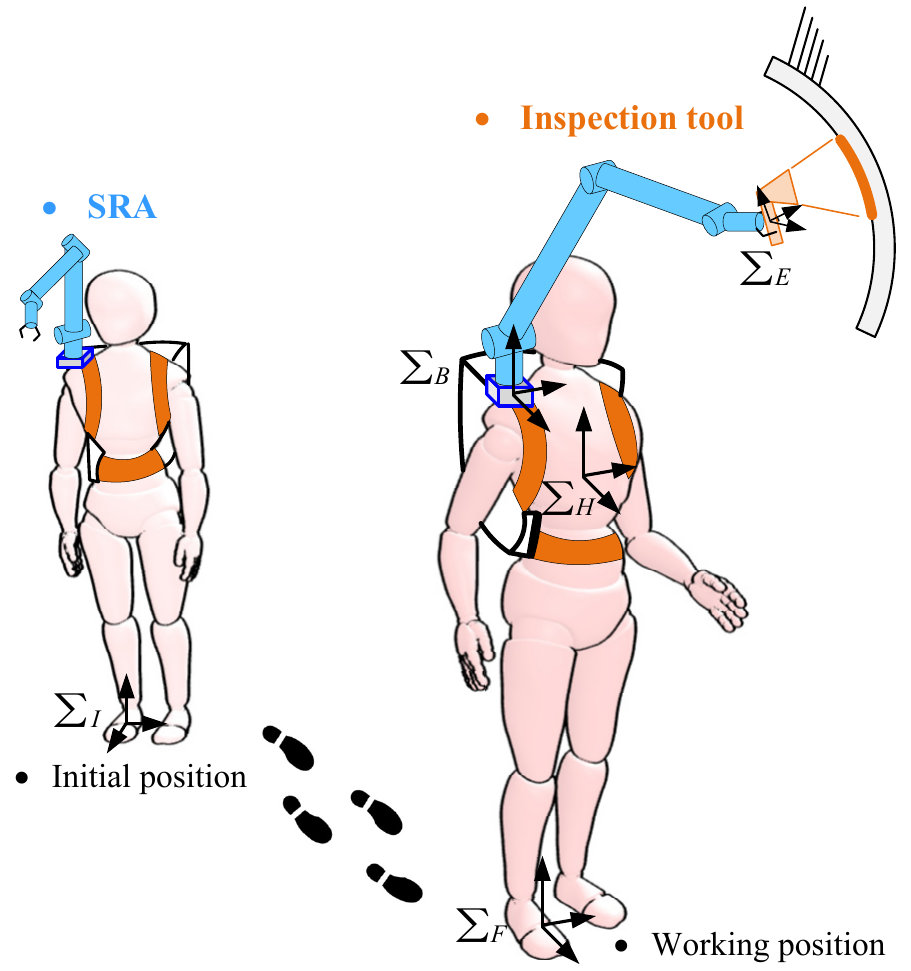}}
\caption{The human-SRA system from the initial calibration position to a local working position for an inspection task.}
\label{fig:inspection}
\end{figure}

Here we briefly introduce the general layout of our SRA. The detailed description is provided in our previous work \cite{Du2023Bi}. The 6-DoF robotic arm is mounted on the user's shoulder, which aims to guarantee the maximum collision-free workspace between the SRA and the human arm. The unique advantage of our SRA is the body adaptability. We design an adjustable mechanical interface allowing different users to manually adjust the installation position of the manipulator to fit different users' body dimensions and switch left/right shoulders. The soft shoulder support keeps the SRA stable and comfortable on the user's body.

\subsection{IMU-Based Sensory Interface}
To achieve the kinematic chain integration of the SRA and the human, the IMU-based motion capture system (Xsens MVN) is chosen as the sensory interface due to its portability. This motion capture system combines the biomechanical model of the human user and an IMU sensor network to estimate the 3D kinematic data of the user. First, the user body dimensions should be measured, including body height, foot size, arm span, ankle height, hip height, hip width, knee height, shoulder width, and shoe sole height. Then, the user performs a calibration process wearing a flexible suit and body straps to attach 17 IMU wireless tracker modules, and the IMU signal data is transmitted to the master (the transmission station and the USB dongle). Through our Xsens-to-ROS bridge \cite{leonori2023bridge}, the user's motion is detected in real time, which works as an important feedback for controlling the SRA.
\section{KINEMATICS MODELING AND ANALYSIS}\label{theory}
As shown in Fig.~\ref{fig:inspection}, the integration system is a floating-base system (with the human user as its base), where $\Sigma _{I}$ is the inertial frame at the initial calibration position, $\Sigma _{F}$ is the right foot frame at the local working position, $\Sigma _{B}$ is the base frame of the manipulator, $\Sigma _{E}$ is the end-effector (EE) frame of the manipulator, and $\Sigma _{H}$ is the human torso frame. As we stated before, we focus on the 3D translation motion compensation control. Two assumed preconditions are given as follows:
\begin{enumerate}
    \item [{(P1)}] The user keeps the orientation of the torso frame $\Sigma _{H}$ stable when moving near the local working position.
    \item [{(P2)}] With the help of the soft shoulder support, $\Sigma _{B}$ and $\Sigma _{H}$ can keep the same orientation and the relative pose remains unchanged when the user is performing translational motions. 
\end{enumerate}

\subsection{Floating-Base Kinematics}
The velocity-level kinematics of the floating-base system\cite{zhang2021attitude} can be established as follows:  
\begin{align}\label{FLK}
\boldsymbol{\dot{x}}_{E}=\boldsymbol{J}_{B}\boldsymbol{\dot{x}}_{B}+\boldsymbol{J}_{M}\boldsymbol{\dot{\theta}_{M}}
\end{align}
where $\boldsymbol{\dot{x}}_{E}=\left [ \boldsymbol{v}_{E}^T, \boldsymbol{\omega}_{E}^T \right ]^T\in \mathbb{R}^{6}$ is the generalized velocity of the EE,  $\boldsymbol{\dot{x}}_{B}=\left [ \boldsymbol{v}_{B}^T, \boldsymbol{\omega}_{B}^T \right ]^T\in \mathbb{R}^{6}$ is the generalized velocity of the human base, $\boldsymbol{\theta}_{M}=\left [ \theta _{1},\theta _{2},\cdots,\theta_{6}\right ]^{T}\in \mathbb{R}^{6}$ is the joint position of the manipulator, and $\boldsymbol{J}_{B}\in \mathbb{R}^{6\times6}$ and $\boldsymbol{J}_{M}\in \mathbb{R}^{6\times6}$ are Jacobian matrices of the base and the manipulator defined in $\Sigma _{I}$. $\boldsymbol{J}_{B}$ is defined as
\begin{equation}
\boldsymbol{J}_{B}=\begin{bmatrix}
\boldsymbol{E}_{3\times3} & -\boldsymbol{\widetilde{p}}_{BE} \\
\boldsymbol{0}_{3\times3} &  \boldsymbol{E}_{3\times3}
\end{bmatrix}
\end{equation}
where $\boldsymbol{E}$ is the identity matrix. The symbol $\left ( \tilde{\cdot}  \right )$ denotes a skew-symmetric matrix, namely, for an arbitrary vector $\boldsymbol{k}=\left [ k_{x},k_{y},k_{z} \right ]^{T}\in \mathbb{R}^{3}$, $\tilde{\boldsymbol{k}}$ is defined as
\begin{align}\label{eq-3}
\tilde{\boldsymbol{k}}=\begin{bmatrix}
0 & -k_{z} &k_{y} \\
k_{z} &0  &-k_{x} \\
 -k_{y}& k_{x} &0
\end{bmatrix}
\end{align}
and $\boldsymbol{p}_{BE}\in \mathbb{R}^{3}$ is the position vector from $\Sigma _{B}$ to $\Sigma _{E}$ in $\Sigma _{I}$:
\begin{equation}
\boldsymbol{p}_{BE}=\boldsymbol{R}_{B}\cdot\boldsymbol{p}_{BE}^{B}
\end{equation}
where $\boldsymbol{R}_{B}\in \mathbb{R}^{3}$ is the orientation of $\Sigma _{B}$ w.r.t. $\Sigma _{I}$, and $\boldsymbol{p}_{BE}^{B}\in \mathbb{R}^{3}$ is the position vector from $\Sigma _{B}$ to $\Sigma _{E}$ in $\Sigma _{B}$. Meanwhile, we can deduce
\begin{equation}
\boldsymbol{J}_{M}=\begin{bmatrix}
\boldsymbol{R}_{B} & \boldsymbol{0}_{3\times3} \\
\boldsymbol{0}_{3\times3} &  \boldsymbol{R}_{B}
\end{bmatrix}\cdot\boldsymbol{J}_{M}^{B}
\end{equation}
where $\boldsymbol{J}_{M}^{B}\in \mathbb{R}^{6\times6}$ is the Jacobian matrix of the manipulator defined in $\Sigma _{B}$. 
According to the assumed precondition (P1), we know $ \boldsymbol{\omega}_{B}\approx\boldsymbol{0}$ and $\boldsymbol{J}_{B}\boldsymbol{\dot{x}}_{B}=[\boldsymbol{v}_{B}^T, \mathbf{0}^T]^T $ when the human user is performing the translational motion at the local working position. Therefore, we can obtain a simplified kinematics model:
\begin{equation}\label{skm}
\begin{bmatrix}
\boldsymbol{v}_{E} \\
\boldsymbol{\omega}_{E}
\end{bmatrix}= \begin{bmatrix}
\boldsymbol{v}_{B} \\
\boldsymbol{0}
\end{bmatrix}+
\begin{bmatrix}
\boldsymbol{J}_{Mv} \\
\boldsymbol{J}_{M\omega}
\end{bmatrix}\boldsymbol{\dot{\theta}_{M}}
\end{equation}
where $\boldsymbol{J}_{Mv}$ and $\boldsymbol{J}_{M\omega}\in \mathbb{R}^{3\times 6}$ are the translational and rotational Jacobian matrices of $\boldsymbol{J}_{M}$, respectively. 

\subsection{Compensation Problem Analysis}

Our basic objective is to keep the EE stable when the user is moving. Due to the translational motion of the human ($\boldsymbol{v}_{B}\neq \boldsymbol{0}$), we compensate for $\boldsymbol{v}_{B}$ by planning $\boldsymbol{\dot{\theta}_{M}}$ to keep $\boldsymbol{x}_{E}$ still (or to follow a desired trajectory). As our manipulator does not have redundant DoFs, it's hard to keep the 6D pose of the EE unchanged at any point in the whole workspace. Since this problem cannot be addressed, we make a compromise by prioritizing subtasks where we assign the EE position compensation ($\boldsymbol{v}_{E}= 0$) as the primary task and the EE orientation compensation ($\boldsymbol{\omega}_{E}= 0$) as the secondary task. According to (\ref{skm}), the primary task of the compensation problem is formulated as
\begin{align}
\label{eq:first_task}%
\underset{\boldsymbol{\dot{\theta}}_{M}}{\text{min}} ~~~ &\|\boldsymbol{J}_{Mv} \boldsymbol{\dot{\theta}}_{M}+\boldsymbol{v}_{B}  \|^2    
\end{align}
Moreover, we assume $\boldsymbol{\dot{\theta}}_{M}^{\star}\in \mathbb{R}^{6}$ is the solution of (7) and the secondary task is to meet
\begin{subequations}\label{eq:second_task}%
\begin{alignat}{2}%
\underset{\boldsymbol{\dot{\theta}}_{M}}{\text{min}} ~~~ &\|\boldsymbol{J}_{M\omega} \boldsymbol{\dot{\theta}}_{M}  \|^2\label{eq:min_goal2}\\ 
\text{s.t} ~~~~ & \boldsymbol{J}_{Mv} \boldsymbol{\dot{\theta}}_{M}=\boldsymbol{J}_{Mv} \boldsymbol{\dot{\theta}}_{M}^{\star}
\end{alignat} 
\end{subequations} 
which means the solution of (8) is within the set of optimizing the primary task \cite{kanoun2012real}. Meanwhile, the planned trajectories should meet the physical constraints ($ \boldsymbol{\theta}_{min} \leq \boldsymbol{\theta}_{M} \leq \boldsymbol{\theta}_{max}$ and $ \boldsymbol{\dot{\theta}}_{min} \leq \boldsymbol{\dot{\theta}}_{M} \leq \boldsymbol{\dot{\theta}}_{max}$), where $\boldsymbol{\theta}_{min}$ and $\boldsymbol{\theta}_{max}\in \mathbb{R}^{6}$ are the lower and upper boundaries of the joint position vector; $\boldsymbol{\dot{\theta}}_{min}$ and $\boldsymbol{\dot{\theta}}_{max}\in \mathbb{R}^{6}$ are the lower and upper boundaries of the joint velocity vector.

\subsection{Inverse Kinematics Solutions}
In this subsection, two approaches are used to solve the above problem: 1) the nullspace-based method (NBM) and 2) the reconstructed Jacobian method (RJM). 
\subsubsection{NBM}
It is one of the most classical approaches to dealing with the task-priority planning problem. This method projects the secondary task into the null space of the Jacobian of the primary task and generates an internal motion without affecting the primary task. In our case, the NBM is formulated as
\begin{gather}
\boldsymbol{\dot{\theta}}_{M} = \boldsymbol{J}_{Mv}^{\dagger} (-\boldsymbol{v}_{B}) +    \boldsymbol{N}\boldsymbol{\xi}_{0}\\
\boldsymbol{N}=\boldsymbol{E}_{6}-\boldsymbol{J}_{Mv}^{\dagger} \boldsymbol{J}_{Mv}
\end{gather}
where for any $\boldsymbol{J}\in \mathbb{R}^{m\times n}$ ($m$$<$$n$), $\boldsymbol{J}^{\dagger }=\boldsymbol{J}^T(\boldsymbol{J} \boldsymbol{J}^{T})^{-1}$ is the pseudo-inverse of $\boldsymbol{J}$, namely the Moore-Penrose pseudo inverse, $\boldsymbol{N}$ is the nullspace projection matrix of $\boldsymbol{J}_{Mv}$, and $\boldsymbol{\xi}_{0}\in \mathbb{R}^{6}$ is an arbitrary vector which is the key to incorporating the secondary task. Here we use the quaternion representation to describe the orientation, namely $\boldsymbol{Q}=\left \{ \eta, \boldsymbol{\epsilon} \right \}$ where $\eta$ is the scalar part and $\boldsymbol{\epsilon}=\left [ \epsilon_{x}, \epsilon_{y},\epsilon_{z} \right ]^T\in \mathbb{R}^{3}$ is the vector part. Let $\boldsymbol{Q}_{d}$ and $\boldsymbol{Q}_{c}$ represent the desired and current orientation, and the orientation error is expressed as 
\begin{equation}\Delta\boldsymbol{Q}=\boldsymbol{Q}_{d}\otimes \boldsymbol{Q}_{c}^{-1}
\end{equation}
where $\Delta\boldsymbol{Q}=\left \{ \Delta\eta, \Delta\boldsymbol{\epsilon} \right \}$ and $\otimes$ is the quaternion product symbol \cite{siciliano2010robotics}. We use the vector part of the quaternion error $\Delta\boldsymbol{\epsilon}$ \cite{yuan1988closed} to design $\boldsymbol{\xi}_{0}$, as 
\begin{equation}
\boldsymbol{\xi}_{0}=\boldsymbol{J}_{M\omega}^{\dagger }\boldsymbol{K}_{O}\Delta\boldsymbol{\epsilon}
\end{equation}
where $\boldsymbol{K}_{O}=diag(K_{O1},K_{O2},K_{O3})\in \mathbb{R}^{3\times 3}$ is the diagonal gain matrix for the EE orientation. 
\subsubsection{RJM} 
Considering the unsolvable problem of any EE pose, we proposed a simple method by reconstructing the Jacobian matrix. As we know,  $\boldsymbol{J}_{M}$ is a ${{6\times6}}$ projection matrix. Our idea is to release some DoFs on the EE orientation and reconstruct a new Jacobian matrix. According to (\ref{skm}), we can obtain
\begin{equation}
\begin{bmatrix}
{\omega}_{Ex} \\
{\omega}_{Ey} \\
{\omega}_{Ez}
\end{bmatrix}=
\begin{bmatrix}
{J}_{M\omega x} \\
{J}_{M\omega y} \\
{J}_{M\omega z}
\end{bmatrix} \boldsymbol{\dot{\theta}}_{M}
\end{equation}
where ${J}_{M\omega}\in \mathbb{R}^{1\times6}$ is the sub-Jacobian matrix for orientation. The new Jacobian matrix $\boldsymbol{J}_{MR}$ is reconstructed as
\begin{equation}
\begin{bmatrix}
\boldsymbol{v}_{E}-\boldsymbol{v}_{B} \\
{\omega}_{Ey} \\
{\omega}_{Ez}
\end{bmatrix}=\underset{\boldsymbol{\boldsymbol{J}_{MR}}}{\underbrace{
\begin{bmatrix}
\boldsymbol{J}_{Mv} \\
{J}_{M\omega y} \\
{J}_{M\omega z}
\end{bmatrix}}}\boldsymbol{\dot{\theta}}_{M}
\end{equation}
where $\boldsymbol{J}_{MR}\in \mathbb{R}^{5\times6}$ and our objective is keeping $\boldsymbol{v}_{E}=\boldsymbol{0}$, ${\omega}_{Ey}=0$, and ${\omega}_{Ez}=0$. According to the closed-loop inverse kinematics algorithm \cite{siciliano2010robotics}, we can deduce
\begin{figure*}[htb]
\centerline{\includegraphics[width=0.85\textwidth]{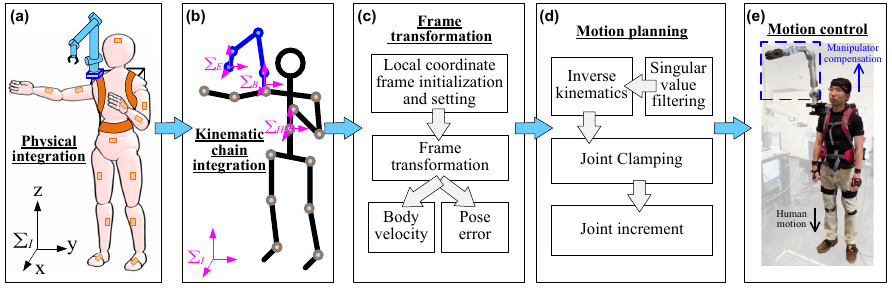}}
\caption{The motion compensation control framework. (a) The physical integration. (b) The kinematic chain integration. (c) The frame transformation. (d) The motion planning. (e) The motion control.} 
\label{fig:framework} 
\end{figure*} 

\begin{equation}
\boldsymbol{\dot{\theta}}_{M}=\boldsymbol{J}_{MR}^{\dagger }\begin{bmatrix}
-\boldsymbol{v}_{B}+\boldsymbol{K}_{P}\Delta{\boldsymbol{p}_{E}}\\ 
{K}_{Oy}\Delta{\epsilon}_{y}\\ 
{K}_{Oz}\Delta{\epsilon}_{z}
\end{bmatrix}
\end{equation}
where $\boldsymbol{K}_{P}=diag(K_{P1},K_{P2},K_{P3})\in \mathbb{R}^{3\times 3}$ is the diagonal gain matrix for the EE position and $\Delta{\boldsymbol{p}_{E}\in \mathbb{R}^{3}}$ is the position variation of the EE.

\section{Motion Compensation Control Framework}\label{framework}
In this section, we introduce our compensation framework. As we aim to plan feasible and safe joint trajectories for real-time control, the singularity handling and the joint limitation are considered in the proposed framework.

\subsection{Compensation Control Framework}
The human motion compensation framework is given in Fig.~\ref{fig:framework}, which includes 5 parts: (a) the physical integration, (b) the kinematic chain integration, (c) the coordinate frame transformation, (d) the motion planning, and (e) the motion control. First, the human user wears the SRA with the IMU-based sensory interface to establish the human-robot integration system. After the calibration, the inertial frame $\Sigma _{I}$ lies in the  right foot of the user at the initial calibration position  (Fig.~\ref{fig:inspection}) and all the kinematic information of the human user is defined w.r.t. $\Sigma _{I}$. As the user usually works at a local working position, the compensation motion control should be finished at the local position. Therefore, we set a local home frame and the human motion should be defined w.r.t. the local home frame. 

\subsection{Coordinate Frame Transformation}

As shown in Fig.~\ref{fig:framework}b, we aim to achieve the motion compensation at a local working position, the human torso frame $\Sigma _{H}$ at $t_0$ (noted as $\Sigma _{H,t_{0}}$) is selected as the local home frame. Therefore, the human translation motion should be defined w.r.t. $\Sigma _{H,t_{0}}$. $\Sigma _{H,t_{0}}$ is set as follows: First, the human user reaches the local working position and stands firmly. Then, we record the local torso frame $\Sigma _{H,t_{0}}$ with its current position $\boldsymbol{p}_{H,t_{0}}$ and its current orientation $\boldsymbol{R}_{H,t_{0}}^{I}$ w.r.t. $\Sigma _{I}$, respectively. According to the assumed precondition (P1), we can deduce the position variation of the local home frame $\Sigma _{H}$ from $t_{0}$ to $t_{k}$ w.r.t. $\Sigma _{H,t_{0}}$ as
\begin{equation}
\Delta\boldsymbol{p}{_{H,t_{k}}^{H,t_{0}}}=\boldsymbol{R}_{I}^{H,t_{0}}(
\boldsymbol{p}_{H,t_{0}} -\boldsymbol{p}_{H,t_{k}})
\end{equation}
where $\boldsymbol{p}_{H,t_{k}}$ is the position of $\Sigma _{H}$ at $t_{k}$  w.r.t. $\Sigma _{I}$. Similarly, we can obtain the local position variation of the base frame $\Sigma _{B}$ from $t_{0}$ to $t_{k}$ w.r.t. $\Sigma _{H,t_{0}}$ as
\begin{equation}
\Delta\boldsymbol{p}{_{B,t_{k}}^{H,t_{0}}}=\boldsymbol{R}_{I}^{H,t_{0}}(
\boldsymbol{p}_{H,t_{0}}+\boldsymbol{p}_{HB,t_{0}}-\boldsymbol{p}_{H,t_{k}}-\boldsymbol{p}_{HB,t_{k}})
\end{equation}
where $\boldsymbol{p}_{HB,t_{0}}$ and $\boldsymbol{p}_{HB,t_{k}}\in \mathbb{R}^{3}$ are the relative position vectors from $\Sigma _{H}$ to $\Sigma _{B}$ at $t_{0}$ and $t_{k}$ w.r.t. $\Sigma _{I}$, respectively. According to the assumed precondition (P2), we know $\boldsymbol{p}_{HB,t_{0}}=\boldsymbol{p}_{HB,t_{k}}$, so we can deduce
\begin{equation}
\Delta\boldsymbol{p}{_{B,t_{k}}^{H,t_{0}}}=\Delta\boldsymbol{p}{_{H,t_{k}}^{H,t_{0}}}
\end{equation}
\begin{equation}
\boldsymbol{v}{_{B,t_{k}}^{H,t_{0}}}=\boldsymbol{v}{_{H,t_{k}}^{H,t_{0}}}=\boldsymbol{R}_{I}^{H,t_{0}}
\boldsymbol{v}_{H,t_{k}}
\end{equation}
where $\boldsymbol{v}_{H,t_{k}}\in \mathbb{R}^{3}$ is the velocity vector of $\Sigma _{H}$ at $t_k$ w.r.t. $\Sigma _{I}$, $\boldsymbol{v}{_{B,t_{k}}^{H,t_{0}}}$ and $\boldsymbol{v}{_{H,t_{k}}^{H,t_{0}}}\in \mathbb{R}^{3}$ are velocity vectors of $\Sigma _{B}$ and $\Sigma _{H}$ at $t_k$ w.r.t. $\Sigma _{H,t_{0}}$, respectively. Moreover, we can deduce the local position variation of $\Sigma _{E}$, namely $\Delta{\boldsymbol{p}_{E}}$ at $t_k$  w.r.t. $\Sigma _{H,t_{0}}$, which will be used to control our SRA:
\begin{equation}
\Delta\boldsymbol{p}{_{E,t_{k}}^{H,t_{0}}}=\Delta\boldsymbol{p}{_{B,t_{k}}^{H,t_{0}}}+\Delta\boldsymbol{p}{_{BE,t_{k}}^{H,t_{0}}}
\end{equation}
where $\Delta\boldsymbol{p}{_{BE,t_{k}}^{H,t_{0}}\in \mathbb{R}^{3}}$ is the relative position from $\Sigma _{B}$ to $\Sigma _{E}$ at $t_k$ w.r.t. $\Sigma _{H,t_{0}}$ and expressed as
\begin{equation} \label{delta-RE}
\Delta\boldsymbol{p}_{{BE,t_{k}}}^{H,t_{0}}=\boldsymbol{R}_{H,t_{k}}^{H,t_{0}}\cdot\boldsymbol{R}_{B,t_{k}}^{H,t_{k}}\cdot 
\boldsymbol{p}_{E,t_{k}}^{B,t_{k}}- \boldsymbol{R}_{B,t_{0}}^{H,t_{0}}\cdot\boldsymbol{p}_{E,t_{0}}^{B,t_{0}}
\end{equation}
where $\boldsymbol{R}_{H,t_{k}}^{H,t_{0}}, \boldsymbol{R}_{B,t_{k}}^{H,t_{k}}$, and $\boldsymbol{R}_{B,t_{0}}^{H,t_{0}}\in \mathbb{R}^{3\times3}$ are the rotation matrices of $\Sigma _{H,t_{k}}$ w.r.t. $\Sigma _{H,t_{0}}$, $\Sigma _{B,t_{k}}$ w.r.t. $\Sigma _{H,t_{k}}$, and  $\Sigma _{B,t_{0}}$ w.r.t. $\Sigma _{H,t_{0}}$, respectively. According to the assumed preconditions (P1) and (P2), each of them is $\boldsymbol{E}_{3\times 3}$ and (\ref{delta-RE}) can be reduced into
\begin{equation}
\Delta\boldsymbol{p}_{{BE,t_{k}}}^{H,t_{0}}=
\boldsymbol{p}_{E,t_{k}}^{B,t_{k}}- \boldsymbol{p}_{E,t_{0}}^{B,t_{0}}
\end{equation}

For the EE orientation stabilization, controlling it w.r.t. $\Sigma _{B}$ is the same as controlling it w.r.t. $\Sigma _{H}$, so we use the following quaternion error to control our SRA:
\begin{equation}\Delta\boldsymbol{Q}_{E,t_{k}}^{B}=\boldsymbol{Q}_{E,t_{0}}^{B,t_{0}}\otimes \boldsymbol{Q}_{E,t_{k}}^{B,t_{k}-1}
=\left \{ \Delta\eta_{E,t_{k}}^{B}, \Delta\boldsymbol{\epsilon}_{E,t_{k}}^{B} \right \}
\end{equation}

\subsection{Singularity Handling}
When the manipulator is close to a singularity configuration, the inverse kinematics becomes unstable and the joint increment becomes large. The singular value filtering (SVF) method \cite{colome2014closed} is adopted in our control framework, which can guarantee the lower-bounded singular values. According to the singular value decomposition, we can deduce
\begin{equation}
\boldsymbol{J}=\boldsymbol{U}\boldsymbol{\Sigma} \boldsymbol{V}^{T}= \sum_{i=1}^{r}\sigma _{i}\boldsymbol{u}_{i}\boldsymbol{v}_{i}^{T}
\end{equation}
where $\boldsymbol{v}_{i}$ and $\boldsymbol{u}_{i}$ are the input and output singular vectors which are the $i$th columns of $\boldsymbol{U}\in \mathbb{R}^{m\times m}$ and $\boldsymbol{V}\in \mathbb{R}^{n\times n}$, respectively, $\sigma _{i}$ is the singular value meeting $\sigma _{1}\geq \sigma _{2}\geq...\geq\sigma _{r}>0$, with $r\leq$ min$\{m,n\}$ being the rank of $\boldsymbol{J}$. The pseudo-inverse of $\boldsymbol{J}$ is $
\boldsymbol{J}^{\dagger}=\boldsymbol{U}\boldsymbol{\Sigma}^{\dagger} \boldsymbol{V}^{T}= \sum_{i=1}^{r}\frac{1}{\sigma _{i}}\boldsymbol{u}_{i}\boldsymbol{v}_{i}^{T}
$. The general damped least-squares pseudo-inverse is expressed as
\begin{equation}
\boldsymbol{J}^{\dagger}_{D}= \sum_{i=1}^{r}\frac{\sigma _{i}}{\sigma _{i}^2 +\lambda^2 }\boldsymbol{u}_{i}\boldsymbol{v}_{i}^{T}
\end{equation}
where $\lambda$ is the damping factor, but a constant small value for $\lambda$ cannot always guarantee good performance in the workspace. Using the SVF method, the filtered Jacobian matrix is defined as
\begin{gather}
\boldsymbol{J}_{F}= \sum_{i=1}^{r}f(\sigma_{i})\boldsymbol{u}_{i}\boldsymbol{v}_{i}^{T} \\
f(\sigma)=\frac{\sigma^{3}+\upsilon\sigma^{2}+2\sigma+2\sigma_{0}}{\sigma ^2+\upsilon\sigma+2}   
\end{gather}
where $f(\sigma)$ is the filtering function, $\sigma_{0}$ is the minimum value that we want to impose on the singular values of $\boldsymbol{J}$, $\upsilon$ is a shape factor, and the corresponding pseudo-inverse of $\boldsymbol{J}_{F}$ is
\begin{equation}
\boldsymbol{J}^{\dagger}_{F}= \sum_{i=1}^{r}\frac{1}{f(\sigma_{i}) }\boldsymbol{u}_{i}\boldsymbol{v}_{i}^{T}
\end{equation}

\subsection{Joint Clamping}

To avoid joint velocity and angle limits, we use the joint clamping algorithm \cite{raunhardt2007progressive} to constrain the joint increment.
\begin{gather}
    \dot{\theta}_{i}=\begin{cases}
	-\dot{\theta}_{iLimit},&	\dot{\theta}_{i}<-\dot{\theta}_{iLimit}\\
        \dot{\theta}_{iLimit},& \dot{\theta}_{i}> + \dot{\theta}_{iLimit}\\
	\dot{\theta}_{i},&		\text{otherwise.} \\\end{cases} 
\end{gather}
where $\dot{\theta}_{iLimit}$ is the $i$th joint velocity boundary. For the joint position at step $k$, we know $\boldsymbol{{\theta}}_{M}(k+1)=\boldsymbol{{\theta}}_{M}(k)+\boldsymbol{H}\Delta\boldsymbol{{\theta}}_{M}$, where $\Delta\boldsymbol{{\theta}}_{M}=\boldsymbol{\dot{\theta}}_{M}\Delta t $ and $\boldsymbol{H}=diag(H_{1},H_{2},...,H_{6})\in \mathbb{R}^{6\times 6}$ is a diagonal activation matrix,
\begin{gather}
    {H}_{i}=\begin{cases}
	1,&	{\theta}_{iMin}<{\theta}_{i}<{\theta}_{iMax}\\
        0,&		\text{otherwise.} \\\end{cases} 
\end{gather}
where ${\theta}_{iMin}$ and ${\theta}_{iMax}$ are the $i$th joint position minimum and maximum boundaries, respectively. Through the joint clamping algorithm, we can guarantee that the planned joint trajectories do not violate the joint limits.

\section{EXPERIMENTAL VERIFICATIONS} \label{experiment}
\begin{figure*}[htb]
\centerline{\includegraphics[width=0.73\textwidth]{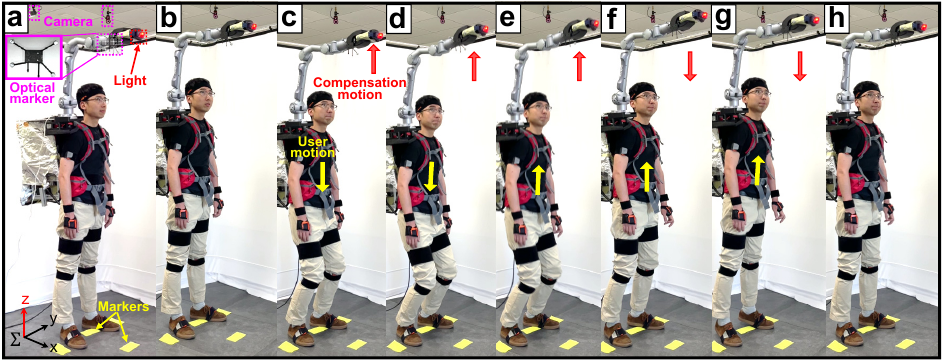}}
\caption{The U-D motion compensation. (a) The optical tracking system. (b) The initial configuration. (c,d) The downward motion. (e-g) The upward motion. (h) The final configuration.} 
\label{fig:U-D} 
\end{figure*}

In this section, we verify our control framework by two sets of experiments: the NBM set and the RJM set. Two types of user motions (1D motion and 3D motion) are investigated in the experiments. The optical tracking motion capture system (in Fig. \ref{fig:U-D}a) is used to obtain the EE position coordinates as absolute coordinates. Although the origin of the optical tracking frame is not the same as the Xsens origin frame $\Sigma _{I}$, its results can be used to evaluate the control performance in Cartesian space. A video of the experiments is available in our YouTube channel\footnote{The video can be found at \url{https://youtu.be/hBFH2gtw55I}}.

\begin{figure*}[htb]
\centerline{\includegraphics[width=0.93\textwidth]{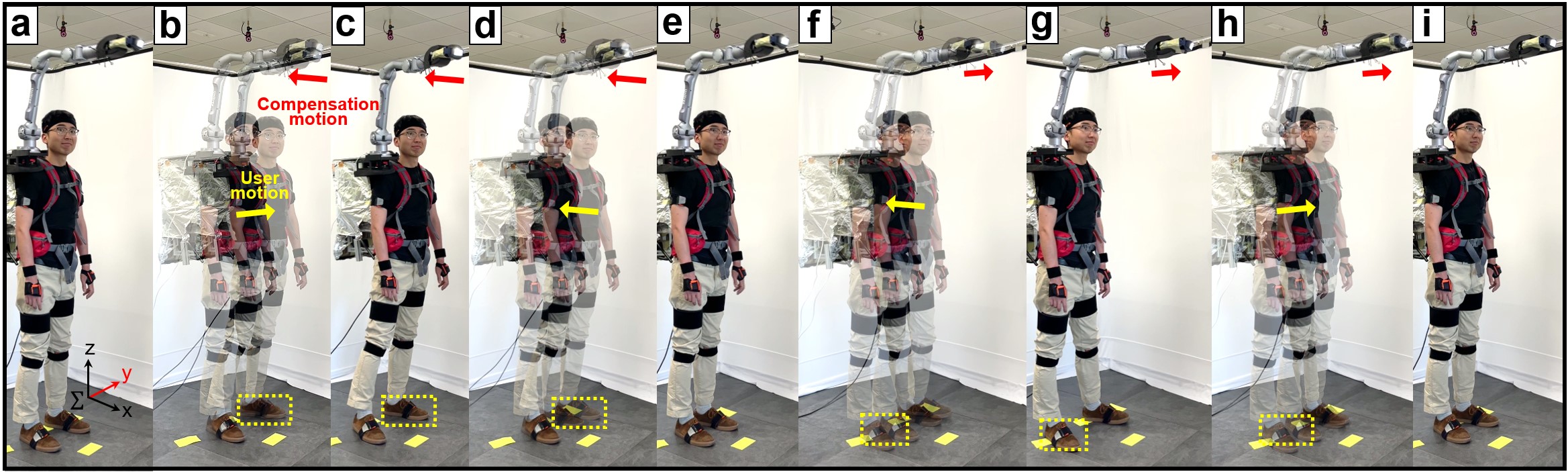}}
\caption{The L-R motion compensation. (a) The initial configuration. (b,c) The left
motion. (d-f) The right motion. (g,h) The second left motion. (i) The final configuration.} 
\label{fig:L-R} 
\end{figure*} 

\begin{figure*}[htb]
\centerline{\includegraphics[width=1.0\textwidth]{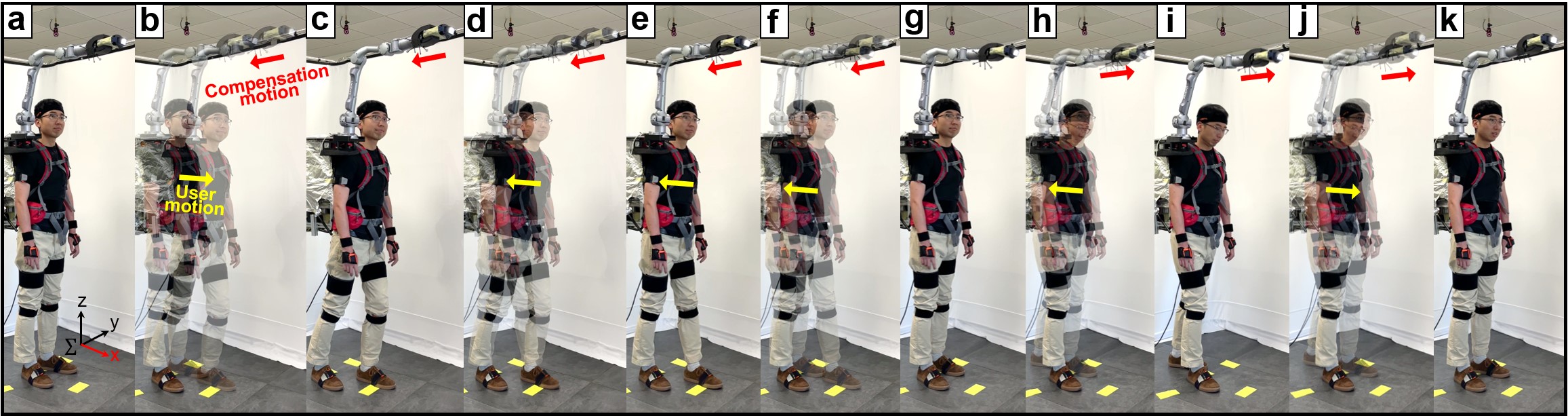}}
\caption{The F-B motion compensation. (a) The initial configuration. (b,c) The forward motion. (d-h) The backward motion. (i,j) The second forward motion. (k) The final configuration.} 
\label{fig:F-B} 
\end{figure*} 

\subsection{Experiments and Results}
\begin{figure*}[htb]
\centerline{\includegraphics[width=1.0\textwidth]{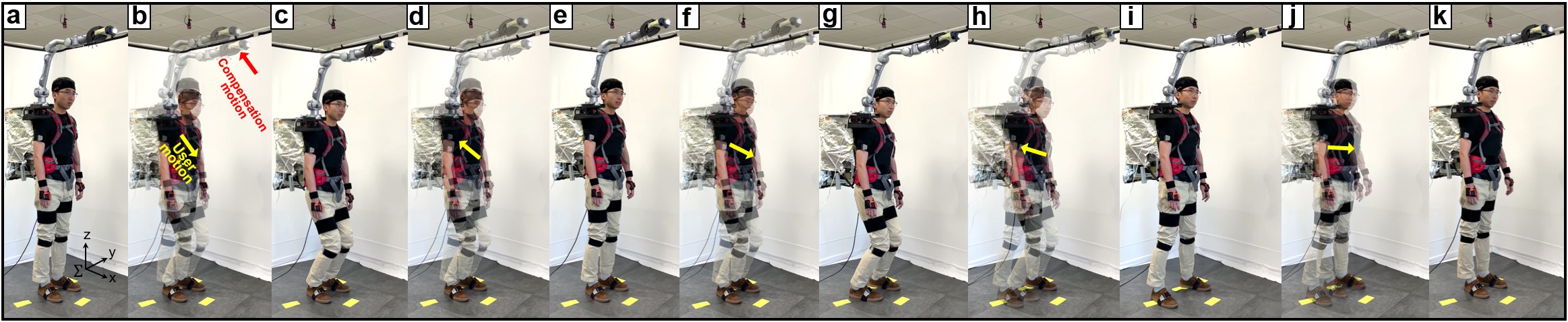}}
\caption{The random motion compensation. (a) The initial configuration. (b,c) The lower left motion. (d,e) The upper right motion. (f,g) The forward left motion. (h,i) The right motion. (j,k) The left motion.} 
\label{fig:Random} 
\end{figure*}

\begin{figure*}[htb]
\centerline{\includegraphics[width=1.0\textwidth]{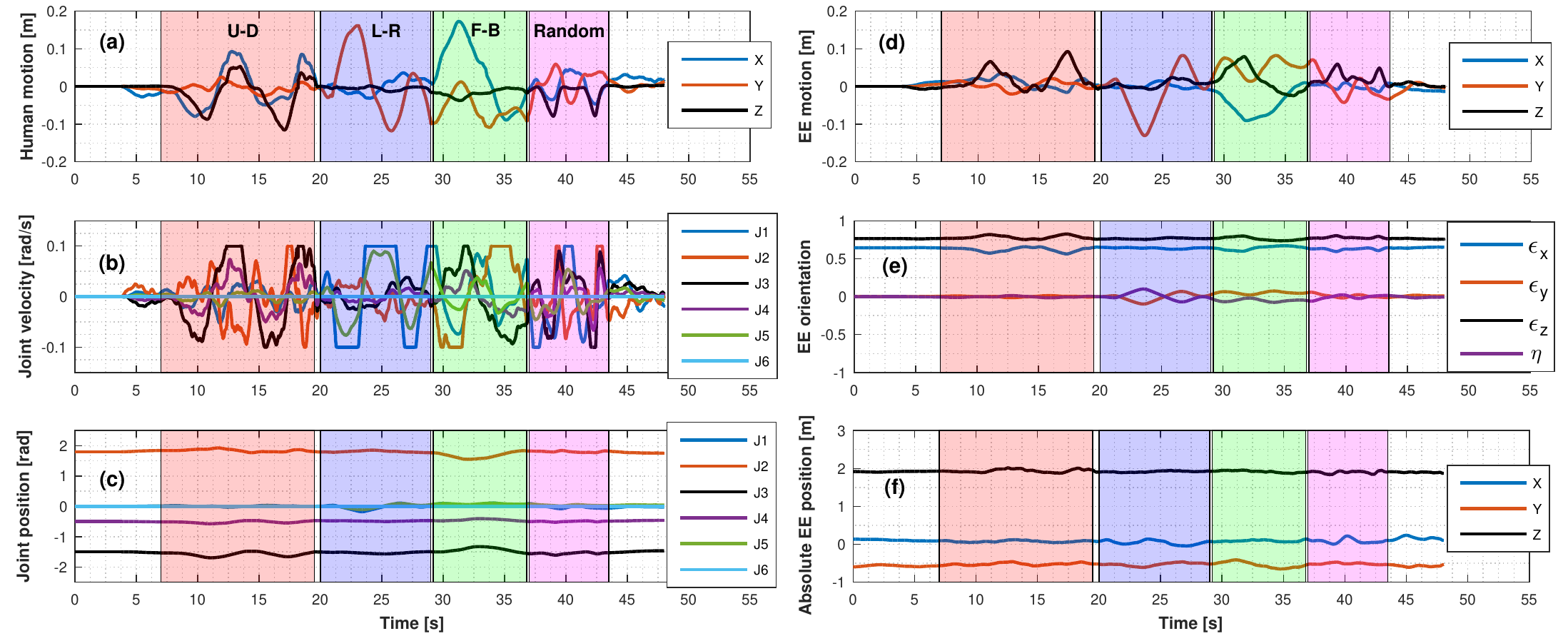}}
\caption{The motion compensation control performance of the NBM set. (a) The human motion w.r.t. $\Sigma _{H,t_{0}}$. (b) The planned joint velocity. (c) The planned joint position. (d) The EE motion w.r.t. $\Sigma _{B,t_{0}}$. (e)The EE orientation w.r.t. $\Sigma _{B,t_{0}}$. (f) The EE position in the optical tracking system.} 
\label{fig:Result-null} 
\end{figure*}

\begin{figure*}[htb]
\centerline{\includegraphics[width=1.0\textwidth]{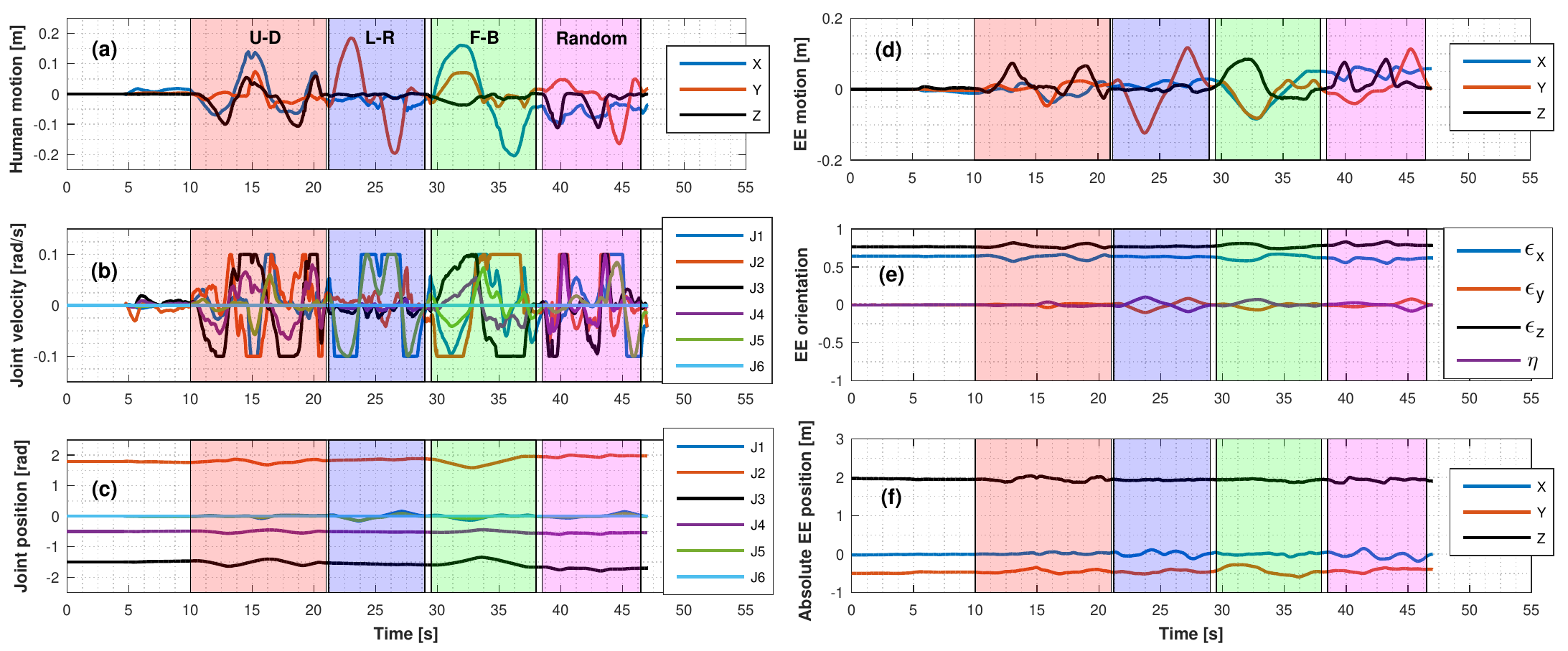}}
\caption{The motion compensation control performance of the RJM set. (a) The human motion w.r.t. $\Sigma _{H,t_{0}}$. (b) The planned joint velocity. (c) The planned joint position. (d) The EE motion w.r.t. $\Sigma _{B,t_{0}}$. (e) The EE orientation w.r.t. $\Sigma _{B,t_{0}}$. (f) The EE position in the optical tracking system.} 
\label{fig:Result-re} 
\end{figure*}

We conduct three 1D motion experiments, namely, the up-down (U-D) motion (Fig.~\ref{fig:U-D}), the left-right (L-R) motion (Fig.~\ref{fig:L-R}), and the forward-backward (F-B) motion (Fig.~\ref{fig:F-B}).  Although the user tries to achieve the 1D motion, the human body's movements are inevitably coupled sometimes. Note that we focus on the z-axis stabilization in the U-D motion, the y-axis stabilization in the L-R motion, and the x-axis stabilization in the F-B motion. As shown in Fig.~\ref{fig:Random}, the human user moves randomly in the 3D motion experiments and we focus on the EE stabilization in 3D space. Accordingly, the experimental results of the NBM and RJM sets are given in Figs. \ref{fig:Result-null} and \ref{fig:Result-re}, respectively. Note that the EE motion is defined w.r.t. $\Sigma _{B,t_{0}}$ and the absolute EE position is defined w.r.t. the origin of the optical tracking system. Moreover, we can see clearly that the z-axis and x-axis are coupled in the U-D motion and the x-axis and y-axis are coupled in the F-B motion. The planned joint velocity trajectories (Figs. \ref{fig:Result-null}b and \ref{fig:Result-re}b) fall in the joint velocity limit $(-0.1, 0.1)$ (rad/s). The EE motion (Figs. \ref{fig:Result-null}d and \ref{fig:Result-re}d) has the opposite trend w.r.t. the human motion (Figs. \ref{fig:Result-null}a and \ref{fig:Result-re}a). Figs. \ref{fig:Result-null}e and \ref{fig:Result-re}e show that EE goes back to its initial orientation after the compensation. Figs. \ref{fig:Result-null}f and \ref{fig:Result-re}f show the compensation performance in Cartesian space.

\begin{figure}[htb]
\centerline{\includegraphics[width=0.42\textwidth]{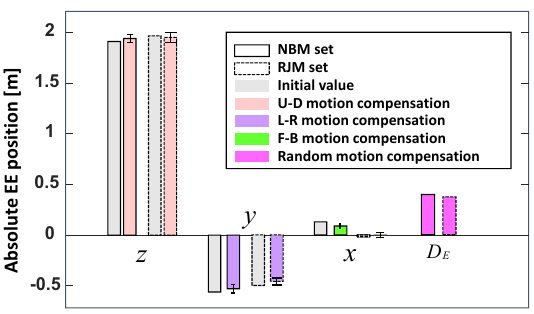}}
\caption{{Evaluation results of two compensation control.}}
\label{fig:statistics}
\end{figure}

\subsection{Evaluation and Analysis }
Two indices are used to evaluate the 1D motion compensation performance.
\begin{gather}    
\label{eq-31}
\bar{ \rho}_{E\chi}=\frac{\sum_{i=k}^{n+k}{ \rho}_{E\chi}\left ( i \right) }{n} \\ s_{\chi} =\sqrt{\frac{\sum_{i=k}^{n+k}\left ( { \rho}_{E\chi}\left ( i \right)-\bar{ \rho}_{E\chi}\right )^{2}}{n-1}}
\end{gather}
where ${\rho}_{E\chi}$ represents the absolute EE position obtained from the optical tracking system, $\bar{\rho}_{E\chi}$ represents the mean value of ${\rho}_{E\chi}$, $\chi$ represents x, y, and z coordinates, and $s_{\chi}$ represents the standard deviation of ${\rho}_{E\chi}(i)$. Meanwhile, we defined ${e}_{\chi}=\left| \bar{\rho}_{E\chi}-\bar{\rho}_{EI\chi}\right|$ to represent the mean EE position error, where $\bar{\rho}_{EI\chi}$ is the initial mean value of ${\rho}_{E\chi}$ $(t\leq 3s)$.

In the NBM set, the initial positions are $\bar{\rho}_{EIx}=0.1280$, $\bar{\rho}_{EIy}=-0.5606$, and $\bar{\rho}_{EIz}=1.9089$. For the U-D motion, $\bar{\rho}_{Ez}=1.9377$, $s_{z}=0.0390$, and ${e}_{z}=0.0288$. For the L-R motion, $\bar{\rho}_{Ey}=-0.5291$, $s_y=0.0411$, and ${e}_{y}=0.0315$. For the F-B motion, $\bar{\rho}_{Ex}=0.0884$, $s_x=0.0191$, and ${e}_{x}=0.0396$.  By comparison, in the RJM set, the initial positions are $\bar{\rho}_{EIx}=-0.0188$, $\bar{\rho}_{EIy}=-0.4993$, and $\bar{\rho}_{EIz}=1.9649$. For the U-D motion, $\bar{\rho}_{Ez}=1.9489$, $s_{z}=0.0497$, and ${e}_{z}=0.0160$. For the L-R motion, $\bar{\rho}_{Ey}=-0.4589$, $s_y=0.0347$, and ${e}_{y}=0.0404$. For the F-B motion, $\bar{\rho}_{Ex}=-0.0024$, $s_x=0.0248$, and ${e}_{x}=0.0164$. Furthermore, we propose a distance-related index to evaluate the 3D motion compensation in Cartesian performance.
\begin{gather}    
\label{eq-33}
 D_E=\frac{\sqrt{e_{x}^2+e_{y}^2+e_{z}^2}}{\sqrt{{e}_{x,max}^2+{e}_{y,max}^2+{e}_{z,max}^2}}
\end{gather} 
where ${e}_{\chi,max}=max(|\bar{ \rho}_{E\chi}-{ \rho}_{E\chi}\left ( i \right) |)$. In the 3D random motion, $D_E$ of the NBE set is $0.3939$ and  $D_E$ of the RJM set is $0.3709$. Meanwhile, the evaluation results ($\bar{ \rho}_{E\chi}$, $s_{\chi}$, and $D_E$) of two experiments are illustrated in Fig. \ref{fig:statistics}.

\subsection{Discussion}
From the experimental results, we can see that the proposed control framework can effectively compensate for the human translation motion. {However, the compensation motion sometimes falls behind the user motion due to the joint clamping (for safety) and the sampling frequency of the IMU-based sensory interface (60 Hz). For example, if the previous step compensation hasn't been finished, the motion compensation in Fig.\ref{fig:F-B} (d-g) still goes backward.} From the evaluation results, we can see that both methods have the same order of magnitude in ${e}_{\chi}$ and $D_E$ (RJM is slightly better than NBM yet simpler). 

\section{CONCLUSIONS AND FUTURE WORK}\label{conclusion}

The motion compensation problem of the human-robot integration system was investigated in this work.
As a special class of floating-base robots, a simplified floating-base kinematics model was deduced to analyze the kinematics of the integration system. As our SRA only has 6 DoFs, which is insufficient to keep the 6D pose of the EE stable, we reformulated the compensation problem by prioritizing the EE position and orientation as the primary task and the secondary task. By virtue of our sensory interfaces, we proposed a motion compensation control framework and a RJM method to solve the task-priority planning problem. Under the proposed framework, the EE position was stabilized in a small range with the mean error ($<$5$cm$). Compared with NBM, the proposed RJM can achieve a competitive control performance with a simpler algorithm structure.

Currently, we do not compensate for the rotational motion of users. In the future, the rotational motion and the users' intention detection will be studied in our compensation framework.



\bibliographystyle{IEEEtran}
\bibliography{manuscript-R1}{}

\end{document}